\documentclass[journal]{IEEEtran}
\usepackage[dvips]{graphicx}
\usepackage{epsfig,multirow}
\usepackage{amsmath,amssymb,bm}
\usepackage{amsfonts,dsfont,color,bbm}
\usepackage{algorithm,algorithmic}
\usepackage[font=footnotesize,labelfont=bf]{caption}
\usepackage{cite,epstopdf,epsf,epsfig, url}

\newtheorem{remark}{\textbf{Remark}}

\renewcommand{\vec}[1]{\mbox{\boldmath${#1}$}}
\newcommand{\norm}[1]{\left|\left|#1\right|\right|}

\begin{document}
\renewcommand{\thepage}{}

\title{A Tree Architecture of LSTM Networks for Sequential Regression with Missing Data}
\author{S. Onur Sahin and Suleyman S. Kozat, \textit{Senior Member, IEEE}
\thanks{
This works is in part supported by Turkish Academy of Sciences Outstanding Researcher Programme and TUBITAK Project No: 117E153.

S. O. Sahin is with the ASELSAN Research Center, Ankara 06370, Turkey, and also with the Department of Electrical and Electronics Engineering,
Bilkent University, Bilkent, Ankara 06800, Turkey (contact e-mail: ssahin@ee.bilkent.edu.tr).

S. S. Kozat is with the Department of Electrical and Electronics Engineering,
Bilkent University, Bilkent, Ankara 06800, Turkey, Tel: +90 (312)
290-2336, Fax: +90 (312) 290-1223, and also with DataBoss A.S. as the CTO.  (contact e-mail:
kozat@ee.bilkent.edu.tr, serdar.kozat@data-boss.com.tr)
}
}
\maketitle
\begin{abstract}
We investigate regression for variable length sequential data containing missing samples and introduce a novel tree architecture based on the Long Short-Term Memory (LSTM) networks.
In our architecture, we employ a variable number of LSTM networks, which use only the existing inputs in the sequence, in a tree-like architecture without any statistical assumptions or imputations on the missing data, unlike all the previous approaches. 
In particular, we incorporate the missingness information by selecting a subset of these LSTM networks based on "presence-pattern" of a certain number of previous inputs.
From the mixture of experts perspective, we train different LSTM networks as our experts for various missingness patterns and then combine their outputs to generate the final prediction.
We also provide the computational complexity analysis of the proposed architecture, which is in the same order of the complexity of the conventional LSTM architectures for the sequence length.
Our method can be readily extended to similar structures such as GRUs, RNNs as remarked in the paper. 
In the experiments, we achieve significant performance improvements with respect to the state-of-the-art methods for the well-known financial and real life datasets.

\end{abstract}
\section{Introduction}\label{sec:intro}
\subsection{Preliminaries} \label{Preliminary}
We study regression of variable length sequential data containing missing samples. Here, we sequentially receive a data sequence suffering from missing input values and estimate an unknown desired signal related to this data sequence. 
In most regression tasks involving sequential data, one usually assumes that we have the complete data sequence \cite{kang2013prevention}. 
However, nearly in every real life application, the data sequences usually contain missing input values due to various reasons such as inconvenience, anomalies and cost savings \cite{de2018nonlinear}, \cite{che2018recurrent}. 
Furthermore, in many real life problems such as medical imaging applications \cite{benedetto2000nonuniform_medimag1} and finance \cite{eng2007algorithms}, we encounter nonuniformly sampled data, which can be modelled as a missing data case \cite{babu2010spectral}. 

To mitigate these issues, the widely used approaches make certain statistical assumptions on the missing data \cite{allison2001missing}, \cite{briggs2003missing}, however, these assumptions usually do not hold and the performance severally degrades in these situations, if the assumptions to not hold \cite{kang2013prevention}.
In our framework, \textbf{we have no such artificial statistical assumptions on the missing data}.
Therefore, our algorithm is less prone to statistical mismatches (if any) and provides a more stable and robust performance in different applications as demonstrated in our simulations.
Specifically, we study the regression problem for variable length data sequences, which contain missing samples, in a supervised framework. In particular, we sequentially observe a data sequence along with its corresponding labels and find a nonlinear relation to predict the labels of the future observations. 

Regression (or prediction as the special case) is extensively studied in the machine learning \cite{lundkvist2014decision}, \cite{you2014multiobjective}, \cite{dufrenois2013formulating} 
and neural network literatures \cite{che2018recurrent}, \cite{weninger2014line}, \cite{xia2018robust}.
The neural network based regression methods are usually preferred in real life applications due to their capability of modelling highly complex and nonlinear structures \cite{greff2017lstm}. 
Among various types of neural networks, in particular, recurrent neural networks (RNNs) are used to process sequences since these networks have an inherent memory storing the past information \cite{greff2017lstm}. 
Although simple RNNs are able to learn temporal behaviour and identify sequential patterns thanks to their memory, they are usually incapable of capturing the long term dependencies due to vanishing and exploding gradient problems \cite{hochreiter1997long}. To resolve these problems, the long short-term memory (LSTM) neural networks \cite{hochreiter1997long}, which are gated RNN architectures with several control structures, are introduced. 

The LSTM networks show a significant performance improvement in sequential data processing applications thanks to their control structures \cite{greff2017lstm}. 
However, this performance usually decreases in real life applications involving missing data \cite{che2018recurrent}, \cite{lipton2016directly}. 
To resolve this issue, one can use imputation techniques for the missing samples and extend the feature vector with an indicator representing whether the corresponding input exists or not, e.g.,  \cite{lipton2015learning}, \cite{lipton2016directly}. 
However, these approaches usually suffer since 
$(i)$ either the imputations indicate only the non-presence of the input and do not contain "any information" from the input sequence or
$(ii)$ the imputations are not adaptable, i.e., they invariably use the same pattern to substitute the missing values.
Hence, they provide less than adequate performance in real life applications \cite{che2018recurrent}.

In this paper, we resolve these problems by introducing a sequential and hierarchical nonlinear learning algorithm based on the LSTM networks, where the outputs of a variable number of LSTM networks are adaptively combined in a tree-like architecture. 
Particularly, our architecture grants each LSTM network the capability of modelling an input sequence with a specific "missingness pattern" (as explained later in the text) and learns to adaptively combine the outputs of these LSTM networks. 
By this way, the proposed algorithm incorporates the missingness information by selecting the particular LSTM networks based on the existence of the certain input patterns.
Hence, it exploits both the input signal itself as well as the missingness pattern to mitigate the effects of the missing samples \textbf{without making any statistical or artificial assumptions on the underlying data}. 
In addition, our architecture keeps the computational load in terms of number of multiplication operations less than the computational load of the conventional algorithms especially when the number of missing samples is high. 
Through an extensive set of experiments, we illustrate significant performance gains compared to the state-of-the-art methods in several real life regression tasks.

\subsection{Prior Art and Comparisons} \label{Prior}

	Among few proposed solutions for processing sequential data containing missing samples, \cite{lipton2015learning} imputes all-zero vectors for these missing inputs. However, this arbitrary input causes deterioration in the information stored in the memory of the LSTM network since the cell and the recurrent input are calculated based on this artificial and unrelated to the input substitutions \cite{che2018recurrent}. On the other hand, in our architecture the LSTM networks process only the inputs included in the data sequence based on the missingness information. Therefore, while we incorporate the missingness information, we completely preserve the content and avoid artifacts due to arbitrary data inclusions, e.g., zero values, mean values, etc., in the memory of the LSTM networks.
	
	In \cite{lipton2016directly}, the authors use forward-filling algorithm to complete the missing data, i.e., they feed the LSTM network with the previous data when the input is missing. They also extend the input vector by a binary missingness indicator, which shows whether the input vector is originally missing or not. Although their algorithm incorporates the missingness information by adding an indicator to the input vector, the information in the memory is corrupted since the same input is feed to the LSTM network multiple times  \cite{che2018recurrent}. In our architecture, the main LSTM network, which contains the essential memory of the architecture, updates the content in the memory once for each input and prevents the redundant contributions from the inputs.
	
	We emphasize that the conventional LSTM based methods \cite{lipton2015learning}, \cite {lipton2016directly} are inadequate to process sequential data containing missing samples since they suffer from certain obstacle such as deterioration in the information stored in the memory. Here, we employ a novel LSTM network based on a tree architecture, which combines the outputs of a variable number of LSTM networks for sequential regression tasks "without" sacrifice from computational load. Our architecture assigns a unique LSTM network based on the missingness information.  
	
\subsection{Contributions} \label{Contr}
Our contributions are as follows.

\begin{enumerate}
\item We introduce a novel LSTM network based on a tree architecture for processing sequential data containing missing samples. 
Our architecture incorporates the missingness information by selection of particular LSTM networks instead of artificially putting the missingness information into the input vectors unlike the conventional methods \cite {lipton2016directly}, \cite{lipton2015learning}.
\item For the first time in the literature, the LSTM networks learn to model the effect of missing inputs only from the existing inputs without any assumptions or imputations on the missing inputs. Therefore, we effectively mitigate the disturbing effects of missing input samples unlike the conventional methods \cite {lipton2016directly}, \cite{lipton2015learning}.
\item Since the proposed architecture uses only the received data to generate its output, our algorithm prevents the deterioration of the information stored in the memory due to disparate or multiple imputations unlike \cite {lipton2016directly}, \cite{lipton2015learning}.
\item Our architecture can be straightforwardly extended to the similar networks working in sequential manner such as RNN and GRU \cite{chung2014empirical}.
\item Through extensive set of experiments involving financial and real life datasets, we demonstrate significant performance gains achieved by our architecture for real life problems with computational complexity \textbf{in the order of the classical approaches}.
\end{enumerate}

\subsection{Organization of the Paper} \label{Org}
The organization of the paper as follows. We formally define our problem setting in Section \ref{sec:prob}. In Section \ref{sec:ProposedSolution}, we first introduce our architecture combining the LSTM networks for an example case to clarify the framework and then extend it to the generic case. In Section \ref{sec:Sim}, we compare the performance of our architecture with respect to the state-of-the-art architectures. The paper concludes with several remarks in Section \ref{sec:Conc}.

\section{Problem Description}\label{sec:prob}
In this paper, all vectors are column vectors and denoted by boldface
lower case letters. Matrices are represented by the boldface capital letters. $x_k$ denotes the $k^\text{th}$  element of the vector $\vec{x}$.
For a vector $\vec{x}$, $\norm{\vec{x}}_1 = \sum |x_k|$ is the $\ell^1$-norm and $\vec{x}^T$ is the ordinary transpose. $\vec{X}_{t_k,j}$ represents the $j^{\text{th}}$ column of the matrix $\vec{X}_{t_k}$, where $t_k$ is the time stamp. 

We sequentially observe variable length vector sequences $\vec{X} = [\vec{x}_{t_1}, ...,\vec{x}_{t_n}] \in \mathcal{X}$, where $\vec{x}_{t_k} \in \mathbb{R}^m$ is the regression vector and $n$ is the length of the sequence $\vec{X}$. Here, the vector sequence $\vec{X}$ is coming at a constant rate, however, $\vec{X}$ contains missing samples, i.e., certain regression vectors, $\vec{x}_{t_k}$, are missing from the data sequence. 
\textcolor{black}{Note that $\vec{x}_{t_k}$ is either completely received or completely missing, i.e., we do not consider the case only certain entries of $\vec{x}_{t_k}$ are missing.}
The desired output for the regression vector $\vec{x}_{t_k}$ is given by $\vec{d}_{t_k} \in \mathbb{R}^u$ and our goal is to estimate $\vec{d}_{t_k}$ by
\begin{align*}
\hat{\vec{d}}_{t_k} = f_{t_k}(\vec{x}_{t_k}, \ldots, \vec{x}_{t_1}, \vec{d}_{t_{k-1}}, \ldots, \vec{d}_{t_1}),
\end{align*}
where $f_{t_k}(\cdot)$ is a possibly time varying and adaptive nonlinear regression function at time step $t_k$. The estimate $\hat{\vec{d}}_{t_k}$ is a function of the current and past observations.
\textcolor{black}{Note that in certain tasks such as the next value prediction problem $\vec{d}_{t_k}$ is not only the output, but also the next input, i.e., $\vec{d}_{t_k} = \vec{x}_{t_{k+1}}$. 
In such cases, either one of $\vec{x}_{t_k}$ or $\vec{d}_{t_k}$ may be missing, whereas the other one exists.
For the tasks where $\vec{d}_{t_k}$ is only the output of $\vec{x}_{t_k}$, i.e., $\vec{d}_{t_k} \neq \vec{x}_{t_{k+1}}$, either both $\vec{x}_{t_k}$ and $\vec{d}_{t_k}$ are received or both are missing.}
For the input $\vec{x}_{t_k}$, the incurred loss is $l(\vec{d}_{t_k}, \hat{\vec{d}}_{t_k})$ and for the whole vector sequence $\vec{X}$, we suffer $E = \frac{1}{n}\sum_{k=1}^{n}l(\vec{d}_{t_k}, \hat{\vec{d}}_{t_k})$.

Since the data has missing samples, the arrival times of the regression vectors are not regular, i.e., the time intervals between the consecutive regression vectors $\vec{x}_{t_k}$ and $\vec{x}_{t_{k+1}}$ may vary and we denote these arrival intervals by $\Delta t_k$'s,
\begin{align*}
\Delta t_k \triangleq t_{k} - t_{k-1}.
\end{align*}

\begin{figure}[t]
  \centering
  \includegraphics[width=.50\textwidth]{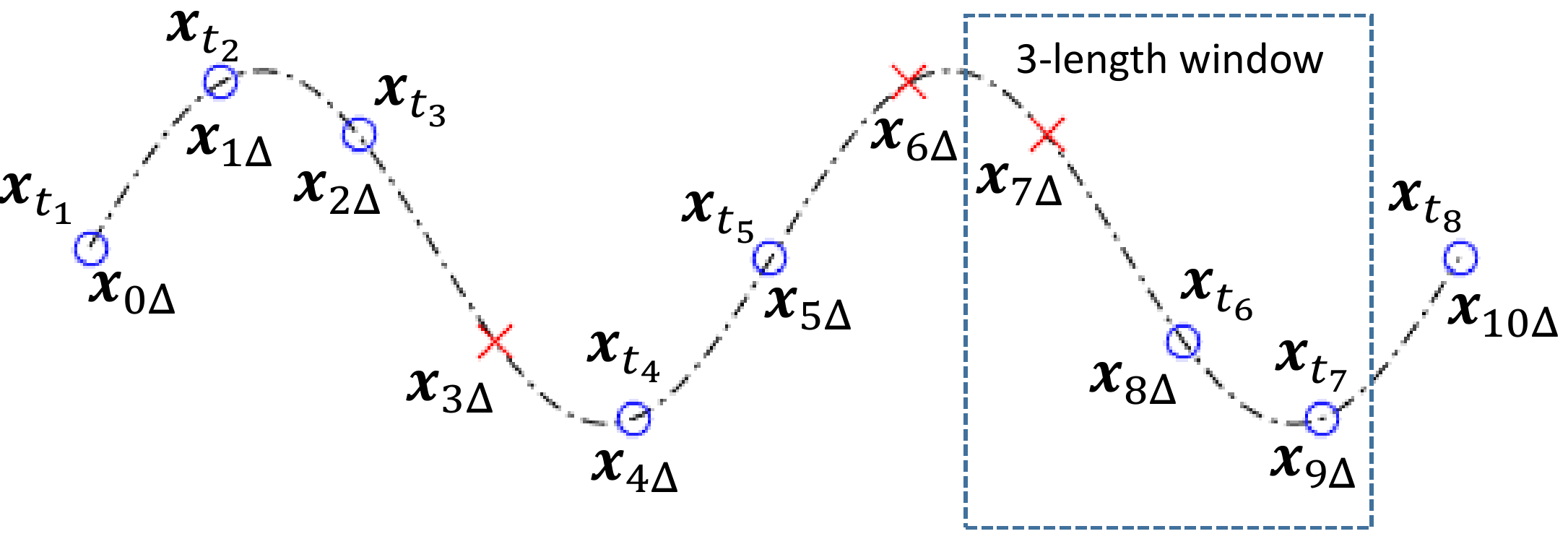}\\
  \caption{An example data sequence with missing inputs.}\label{fig:sine_example}
  \vspace{-0.3cm}
\end{figure}

To clarify the framework, in Fig. \ref{fig:sine_example}, we illustrate an example data sequence $\vec{X} = [x_{t_1}, ...,x_{t_8}]$, where each vector is selected as a scalar, i.e., $u = 1$,  with constant time intervals $\Delta$. However, the data sequence has missing samples, i.e., $x_{3\Delta}$, $x_{6\Delta}$ and $x_{7\Delta}$. Here, $ x_{t_k}$ represents the samples in a received order, e.g., $x_{t_2} = x_\Delta$. $ x_{m\Delta}$ is the data at time $m\Delta$, which we may receive, e.g., $x_{2\Delta}$, or may not receive, e.g.,  $x_{3\Delta}$. 
Hence, for this sequence $x_{t_1} = x_{0\Delta}, x_{t_2}= x_{1\Delta}, x_{t_3}=x_{2\Delta}, x_{t_4}= x_{4\Delta}, x_{t_5}= x_{5\Delta}, x_{t_6}=x_{8\Delta}, x_{t_7}= x_{9\Delta}$ and $x_{t_8}= x_{10\Delta} $.
Since the time intervals between the consecutive regression vectors are not regular, the regression function should adapt different cases to predict the desired signal. 
As an example, let us consider one step ahead prediction as a special regression task for this input sequence. Then, $x_{t_3}=x_{2\Delta}$ should be predicted using the $x_{0\Delta}$ and $x_{1\Delta}$. However, $x_{t_4}=x_{4\Delta}$ should be predicted using the $x_{2\Delta}$, $x_{1\Delta}$ and $x_{0\Delta}$ since $x_{3\Delta}$ is missing. 

Here, we use recurrent neural networks to generate the sequential estimates $\hat{d}_{t_k}$. A basic RNN structure is given by \cite{jaeger2002tutorial}
\begin{equation}
\label{eq:RNN}
\begin{aligned}
\vec{h}_{t_k} &= f(\vec{W}_h\vec{x}_{t_k} + \vec{R}_h \vec{h}_{t_{k-1}})  \\
\vec{y}_{t_k} &= g(\vec{R}_y\vec{h}_{t_k}),
\end{aligned}
\end{equation}
where $\vec{x}_{t_k} \in \mathbb{R}^m$ is the regression vector, $\vec{h}_{t_k} \in \mathbb{R}^q$ is the state vector and $\vec{y}_{t_k} \in \mathbb{R}^q$ is the output at time $t_k$.
$\vec{W}_h \in \mathbb{R}^{q\times m}$, $\vec{R}_h \in \mathbb{R}^{q\times q}$ represent the input weight matrices, $\vec{R}_y \in \mathbb{R}^{q\times q}$ is the output weight matrix. $f(\cdot)$ and $g(\cdot)$ are the nonlinear functions and apply point-wise operations.

\begin{figure}[t]
  \centering
  \includegraphics[width=.49\textwidth]{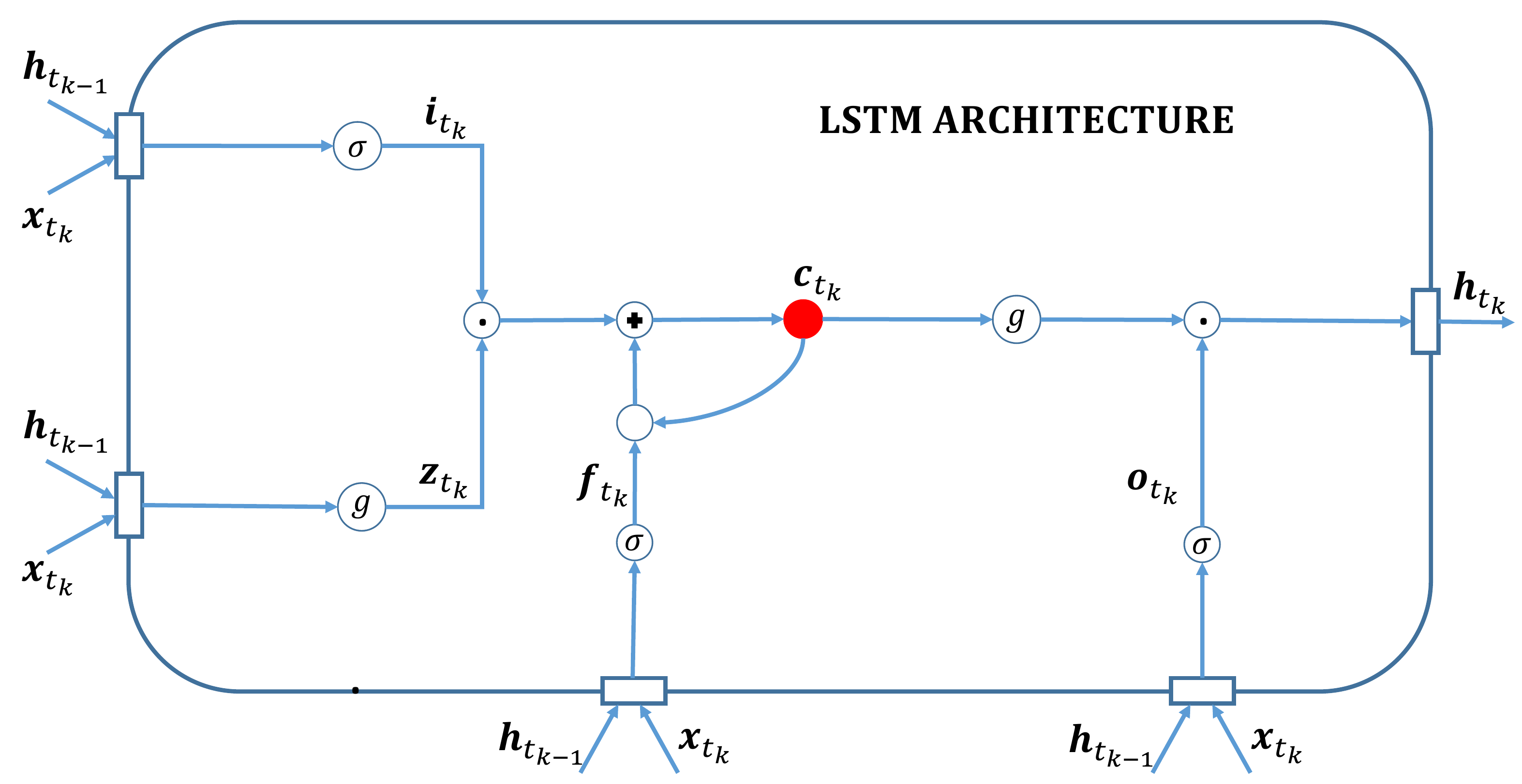}\\
  \caption{Detailed schematic of the LSTM architecture.}\label{fig:LSTM}
  \vspace{-0.3cm}
\end{figure}

As a special case of the RNNs, we focus on the LSTM networks. Among many different variants of the LSTM architecture, we use the most widely used variant, i.e., the LSTM architecture without peephole connections illustrated in Fig. \ref{fig:LSTM}. The LSTM architecture is given by the following set of equations:
\begin{align}
\vec{z}_{t_k} &= g(\vec{W}_z\vec{x}_{t_k} + \vec{R}_z\vec{h}_{t_{k-1}}) \label{eq:zt}\\
\vec{i}_{t_k} &= \sigma(\vec{W}_i\vec{x}_{t_k} + \vec{R}_i\vec{h}_{t_{k-1}})\label{eq:it}\\
\vec{f}_{t_k} &= \sigma(\vec{W}_f\vec{x}_{t_k} + \vec{R}_f\vec{h}_{t_{k-1}})\label{eq:ft}\\
\vec{o}_{t_k} &= \sigma(\vec{W}_o\vec{x}_{t_k} + \vec{R}_o\vec{h}_{t_{k-1}})\label{eq:ot}\\
\vec{c}_{t_k} &= \vec{i}_{t_k} \odot \vec{z}_{t_k} + \vec{f}_{t} \odot \vec{c}_{t_{k-1}} \label{eq:ct}\\
\vec{h}_{t_k} &= \vec{o}_{t_k} \odot g(\vec{c}_{t_k}) \label{eq:ht},
\end{align}
where $\vec{x}_{t_k} \in \mathbb{R}^m$ is the input vector, $\vec{c}_{t_k} \in \mathbb{R}^q$ is the state vector and $\vec{h}_{t_k} \in \mathbb{R}^q$ is the output vector of the LSTM network at time $t_k$. $\vec{z}_{t_k}$ is the block input, $\vec{i}_{t_k}$, $\vec{f}_{t_k}$, $\vec{o}_{t_k} \in \mathbb{R}^q$ represent the input, forget and output gates at time ${t_k}$, respectively.  $\vec{W}_z$, $\vec{W}_i$, $\vec{W}_f$, $\vec{W}_o \in \mathbb{R}^{q\times m}$ are the input weight matrices and $\vec{R}_z$, $\vec{R}_i$, $\vec{R}_f$, $\vec{R}_o \in \mathbb{R}^{q\times q}$ are the recurrent input weight matrices. $g(\cdot)$ and $\sigma(\cdot)$ are the point-wise nonlinear activation functions. $g(\cdot)$ is commonly set to the tangent hyperbolic function, i.e., $\text{tanh}(\cdot)$ and $\sigma(\cdot)$ is the sigmoid function. With the abuse of notation, we incorporate the bias weights, $\vec{b}_z$, $\vec{b}_i$, $\vec{b}_f$, $\vec{b}_o \in \mathbb{R}^{q}$, into the input weight matrices and denote them by $\vec{W}_\theta = [\vec{W}_\theta;\vec{b}_\theta]$, $\theta \in \{z, i, f, o \}$, where $\vec{x}_{t} = [\vec{x}_{t};1]$.
	As described in the following section, we estimate the desired signal $d_{t_k}$ by
\begin{align}
\hat{d}_{t_k} = \vec{\hat{w}}_{t_k}^T\vec{\hat{h}}_{t_k},
\end{align}
where $\vec{\hat{w}}_{t_k} \in \mathbb{R}^q$ is the regression coefficients. To obtain $\vec{\hat{h}}_{t_k}$, we adaptively combine the outputs of the different LSTM networks in our architecture by
\begin{align}
\vec{\hat{h}}_{t_k} = \sum_{i=1}^{K_{t_k}} \alpha_{t_k}^{(i)}\vec{h}_{t_k}^{(i)},
\end{align} 
where $\vec{h}^{(i)}_{t_k}$ is the output of the $i^\text{th}$ LSTM network, $K_{t_k}$ is the number of the total LSTM networks to be combined at time $t_k$ and $\vec{\hat{h}}_{t_k}$ is the linear combination of these outputs.

In the following, we introduce a tree architecture based on the LSTM networks working on the sequential data with missing samples, and also provide its forward-pass formulas.

\begin{figure*}[t]
  \centering
  \includegraphics[width=.75\textwidth]{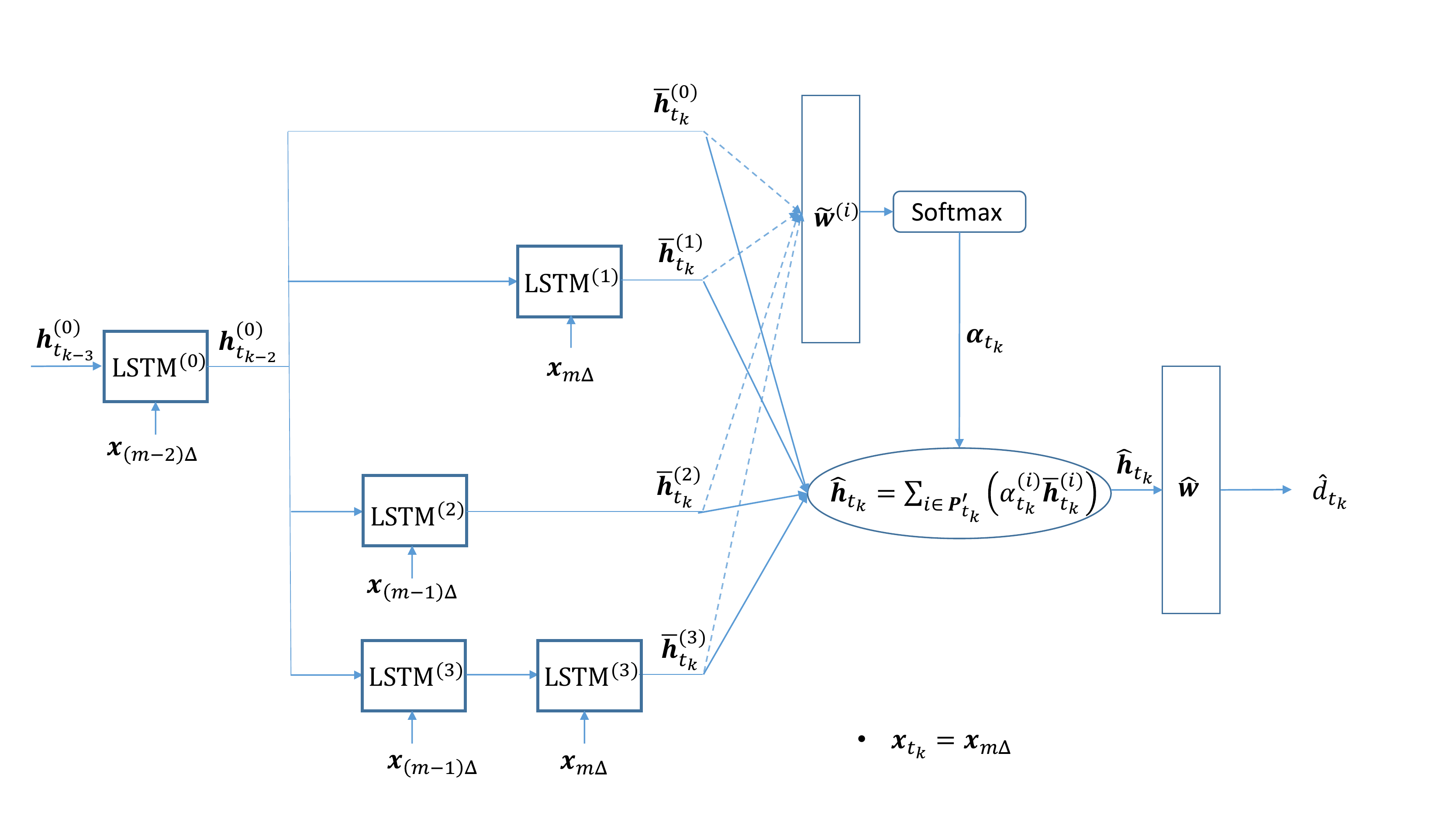}\\
  \caption{Detailed schematic of the Tree-LSTM architecture, where the tree depth $L = 2$. Note that $\vec{x}_{t_k} = \vec{x}_{m\Delta}$.}\label{fig:model}
  \vspace{-0.5cm}
\end{figure*}

\section{LSTM Network Based Tree Architecture}\label{sec:ProposedSolution}
Since the time intervals between the regression vectors are not regular in the case of missing data, the regression function should adapt to different scenarios to estimate the desired signal.
Hence, we directly incorporate the missingness information into our nonlinear regression function $f_{t_k}(\cdot)$ to model the effect of the missing data in our sequence. 
In our algorithm, we consider the missing input values in a particular window, which shifts at each time step. The length of this window is a user defined parameter and adjusts the modeling capacity vs. trainability trade-off. We investigate the effect of window length in Section \ref{sec:Sim}.

For this purpose, we first partition the regression function $f_{t_k}(\cdot)$ into two parts as follows
\begin{align}
\hat{d}_{t_k} =& \theta_{t_k}^{\text{M}} f_{t_k}^{\text{M}}(\cdot) + \theta_{t_k}^{\text{W}} f_{t_k}^{\text{W}}(\cdot), \label{eq:twopart}
\end{align}
where $f_{t_k}^{\text{W}}(\cdot)$ processes the input sequence in a particular window, e.g., the length-3 window in Fig. \ref{fig:sine_example}, and $f_{t_k}^{\text{M}}(\cdot)$ is the main regression function using the whole input sequence except the samples inside the window.
The purpose of two distinct functions will be clear in the following. 

Specifically, for a length-$L$ window, $f_{t_k}^{\text{M}}(\cdot)$ and $f_{t_k}^{\text{W}}(\cdot)$ process the inputs $[\vec{x}_{0\Delta}, \ldots, \vec{x}_{(m-L)\Delta}]$ and $[\vec{x}_{(m-L +1)\Delta}, \ldots, \vec{x}_{m\Delta}]$, where $t_k = m\Delta$, respectively.
Here, $f_{t_k}^{\text{M}}(\cdot)$ captures the general pattern of the data, while $f_{t_k}^{\text{W}}(\cdot)$ provides more elaborate decisions on the inputs inside the window. 
Next, we incorporate the missingness information into $f_{t_k}^{\text{W}}(\cdot)$, i.e., $f_{t_k}^{\text{W}}(\cdot, \vec{p}_{t_k}^{(L)})$. 

We define "presence-pattern", i.e., the pattern of the present input samples,  $\vec{p}^{(L)}_{t_k} = [p^{(L)}_{t_k,1}, \ldots, p^{(L)}_{t_k,L}] \in \{0,1\}^L$, which holds the missingness information in its most explicit form, i.e., whether the inputs $[\vec{x}_{(m-L+1)\Delta}, \ldots, \vec{x}_{m\Delta}]$ exist or not, where $\vec{x}_{t_k} = \vec{x}_{m\Delta}$. 
For example, a length-3 presence-pattern $\vec{p}^{(3)}_{t_k} = [1,0,1]$ indicates that $\vec{x}_{m\Delta}$ and $\vec{x}_{(m-2)\Delta}$ are received, however, $\vec{x}_{(m-1)\Delta}$ is missing from the input sequence, where $t_k = m\Delta$.
The presence-pattern always has the same length with the window, i.e., $L = W$, hence, we drop the length $L$ to simplify the notation, i.e., $\vec{p}_{t_k}$.  
Note that we explicitly incorporate the missingness information into $f_{t_k}^{\text{W}}(\cdot, \vec{p}_{t_k})$, besides, $f_{t_k}^{\text{M}}(\cdot)$ carries this information for the past inputs thanks to the memory in the LSTM architecture. Hence, we also provide this missingness information for the former inputs.

There exist $L$ input vectors inside the window of length $L$, which corresponds to $2^L$ possible unique presence-patterns since each input vector has two options, i.e., may or may not exist. Since all-zero presence-pattern indicates all of the inputs inside the window are missing, we have $2^L - 1$ presence-patterns containing inputs to be processed. 
In our algorithm, we assign a unique regression function to process each of these patterns, while $f_{t_k}^{\text{M}}(\cdot)$ corresponds to all-zero pattern since it only uses the inputs outside the window, which is the main reason using two functions in \eqref{eq:twopart}. For this purpose, we divide $f_{t_k}^{\text{W}}(\cdot, \vec{p}_{t_k})$ into $2^L - 1$ components as follows
\begin{align}
\hat{d}_{t_k} =& \theta_{t_k}^{\text{M}} f_{t_k}^{\text{M}}(\cdot) + \theta_{t_k}^{\text{W}} \sum_{i =1}^{2^L -1} \beta^{(i)}_{t_k} f_{t_k}^{\text{W}_i}(\cdot, \vec{p}_{t_k})\label{eq:partitioned}, 
\end{align}
where $\theta_{t_k}^{\text{M}}$, $\theta_{t_k}^{\text{W}}$ and $\beta^{(i)}_{t_k} \in \mathbb{R}$. Each $f_{t_k}^{\text{W}_i}(\cdot)$ is the regression function specifically assigned to process the input sequence with a unique presence-pattern $\vec{p}^{(i)}$. 
As an example, $f_{t_k}^{\text{W}_1}(\cdot)$, $f_{t_k}^{\text{W}_5}(\cdot)$ and $f_{t_k}^{\text{W}_7}(\cdot)$ process the inputs for length-3 presence-patterns $\vec{p}^{(1)} = [0,0,1]$, $\vec{p}^{(5)} = [1,0,1]$ and $\vec{p}^{(7)} = [1,1,1]$, respectively, which is the binary representation of the number $i$ using $L$ bits.  
We can also consider $f_{t_k}^{\text{M}}(\cdot)$ is the regression function for the all-zero presence-pattern $\vec{p}^{(0)} = [0,0,0]$. 
\textcolor{black}{In its most extensive form, our model contains $2^L$ unique regression functions, the computational loads and the methods for reducing the number of regression functions will be explained in Section  \ref{subsec:Generic}.}

One can directly use $f_{t_k}^{\text{W}_i}(\cdot)$ to process the inputs inside the window, when $\vec{p}^{(i)} = \vec{p}_{t_k}$. 
However, note that certain presence-patterns inherently contain the other presence-patterns, therefore, we can use multiple regression functions to improve the estimate $\hat{d}_{t_k}$ for the same $\vec{p}_{t_k}$. 
For example, when $\vec{p}_{t_k} = [0, 1, 1]$ is received, we also obtain the patterns $[0, 0, 0]$, $[0, 0, 1]$ and $[0, 1, 0]$. 
Therefore, we have sufficient information to use and train $f_{t_k}^{\text{M}}(\cdot)$, $f_{t_k}^{\text{W}_1}(\cdot)$ and $f_{t_k}^{\text{W}_2}(\cdot)$ in addition to $f_{t_k}^{\text{W}_3}(\cdot)$. 
To represent this relation, we define presence subpattern $\bar{\vec{p}}_{t_k}$ such that 
if $\max {(\bar{\vec{p}}_{t_k, j}, \vec{p}_{t_k, j})} = \vec{p}_{t_k, j}, \forall j$ 
excluding $\bar{\vec{p}}_{t_k} = \vec{p}_{t_k}$, then $\bar{\vec{p}}_{t_k}$ is a subpattern of $\vec{p}_{t_k}$.
Next, we define the set $\vec{P}_{t_k}$ as the active set of the presence-pattern $\vec{p}_{t_k}$ such that $\vec{P}_{t_k}$  contains all possible subpatterns of $\vec{p}_{t_k}$ in addition to $\vec{p}_{t_k}$ itself. 
To simplify the notation, we also define the set $\vec{P}^{\prime}_{t_k}$ such that $\vec{P}^{\prime}_{t_k}$ contains the decimal representations of the presence-patterns included in the set $\vec{P}_{t_k}$. 
As an example, for $\vec{p}_{t_k} = [1,0,1]$, the active set $\vec{P}_{t_k}  = \{[0,0,0], [0,0,1], [1,0,0], [1,0,1]\}$ and $\vec{P}^{\prime}_{t_k}  = \{0, 1, 4, 5\}$. 
Note that all-zero pattern, i.e., $[0,0,0]$ for $L = 3$, is always included in the active set.

In our architecture, we use a separate LSTM network to model each regression function in \eqref{eq:partitioned}. 
In particular, we employ one main LSTM network modelling $ f_{t_k}^{\text{M}}(\cdot)$ and also many different leaf LSTM networks without computational increase thanks to our tree approach, which model the regression functions $ f_{t_k}^{\text{W}_i}(\cdot)$ in \eqref{eq:partitioned}. 
While the main LSTM network captures the general pattern of the data, the leaf LSTM networks provide more precise outputs based on the presence-pattern inside the window. 
We pass the state and the output of the main LSTM network to the leaf LSTM networks as their initial states and recurrent inputs to provide them with the information on the history of the sequence.
We then combine the outputs of these LSTM networks to generate our final output.

In our algorithm, each leaf LSTM network is assigned to a particular presence-pattern. 
If an input is missing in a length-$L$ window, the LSTM networks containing this input in their assigned input sequence do not generate their outputs.
Therefore, only a subset of the leaf LSTM networks contribute to the final output based on the existence of the inputs inside the particular window in case of missing data.  
By this way, we directly incorporate the missingness information by selecting the particular leaf LSTM networks instead of artificially inserting it into the input vectors as done in literature \cite{lipton2015learning}, \cite{lipton2016directly}. 
Due to this hierarchical nature we name our architecture as the Tree-LSTM architecture.

To clarify the algorithm, let us say we receive a sequence with missing samples as illustrated in Fig. \ref{fig:sine_example} and the aim is to predict the next sample, i.e., $d_{t_k} = \vec{x}_{t_{k+1}}$. 
For example, to estimate $\vec{x}_{t_8} = \vec{x}_{10\Delta}$ in Fig. \ref{fig:sine_example}, the length-3 window encapsulates the inputs $[\vec{x}_{7\Delta}, \vec{x}_{8\Delta}, \vec{x}_{9\Delta}]$. The main LSTM network processes the existing inputs before this window, i.e., $[\vec{x}_{t_1}, \ldots, \vec{x}_{t_5}] = [\vec{x}_{0\Delta}, \ldots, \vec{x}_{5\Delta}]$ and generates its state and output vectors. Since $\vec{x}_{7\Delta}$ is missing from our sequence, only the leaf LSTM networks, which do not contain $\vec{x}_{7\Delta}$ in their input sequences, are able to generate their outputs. Next, we combine the outputs of different LSTM networks and obtain our final estimate. 

In Section \ref{subsec:Specific}, we first provide our architecture with a specific depth to clarify the framework. In particular, we select the depth as $L=2$ to provide a clear representation of the algorithm with a small number of LSTM networks. We then extend this architecture to the generic case in \ref{subsec:Generic}.
\subsection{A Specific Tree-LSTM Architecture}\label{subsec:Specific}

Suppose the depth of the Tree-LSTM network is $L=2$ and we estimate the generic desired signal as in Fig. \ref{fig:model}.
The architecture contains $2^{2} = 4$ different LSTM networks. 
For each LSTM network, $\vec{W}_z^{(j)}$, $\vec{W}_i^{(j)}$, $\vec{W}_f^{(j)}$, $\vec{W}_o^{(j)} \in \mathbb{R}^{q\times m}$ are the input weight matrices and $\vec{R}_z^{(j)}$, $\vec{R}_i^{(j)}$, $\vec{R}_f^{(j)}$, $\vec{R}_o^{(j)} \in \mathbb{R}^{q\times q}$ are the recurrent weight matrices of the $j^\text{th}$ LSTM network, i.e., $\text{LSTM}^{(j)}$ in Fig. \ref{fig:model}.

This architecture as shown in Fig. \ref{fig:model} contains four different LSTM networks, i.e., $\text{LSTM}^{(0)}$, $\text{LSTM}^{(1)}$, $\text{LSTM}^{(2)}$ and $\text{LSTM}^{(3)}$, where each LSTM network is responsible for processing the data sequence with a particular presence-pattern. In particular, $\text{LSTM}^{(0)}$, $\text{LSTM}^{(1)}$, $\text{LSTM}^{(2)}$ and $\text{LSTM}^{(3)}$ are assigned to the presence-patterns $[0, 0]$, $[0, 1]$, $[1, 0]$ and $[1, 1]$, respectively. Here, we have one main LSTM network, i.e., $\text{LSTM}^{(0)}$, to identify the general pattern and propagate the essential state information contained in the state and output vectors $\vec{c}_{t_k}^{(0)}$ and $\vec{h}_{t_k}^{(0)}$. The other three LSTM networks, i.e., $\text{LSTM}^{(1)}$, $\text{LSTM}^{(2)}$ and $\text{LSTM}^{(3)}$, are the leaf LSTM networks. They receive  $\vec{c}_{t_k}^{(0)}$ and $\vec{h}_{t_k}^{(0)}$ as their initial states and process their input sequences inside the length-2 window. 
Note that while the $\text{LSTM}^{(0)}$ network runs over the whole sequence, the other three LSTM networks process only the data sequence corresponding to their presence-patterns in this window. 

When a regression vector $\vec{x}_{t_k}$ with a particular presence-pattern $\vec{p}_{t_k}$ is received, only the LSTM networks included in the active set of $\vec{p}_{t_k}$, i.e., $\vec{P}^{\prime}_{t_k}$, process  their corresponding input sequences and generate their outputs. 
For example, when $\vec{p}_{t_{k}} = [1 , 1]$, all of the four LSTM networks generate output since $\vec{P}^{\prime} = \{0, 1, 2, 3 \}$.
To generate these outputs, firstly, the main LSTM network processes the input $\vec{x}_{t_{k-2}}$ and generates its state and output vectors, $\vec{c}_{t_{k-2}}^{(0)}$ and $\vec{h}_{t_{k-2}}^{(0)}$, respectively. As shown in Fig. \ref{fig:model}, these vectors are passed to the leaf LSTM networks as their initial states and recurrent inputs.
Then, each leaf LSTM network processes its corresponding input sequence, i.e., the input is merely $\vec{x}_{t_k}$ for $\text{LSTM}^{(1)}$, similarly, merely $\vec{x}_{t_{k-1}}$ for $\text{LSTM}^{(2)}$ and $[\vec{x}_{t_{k-1}}, \vec{x}_{t_{k}}]$ for $\text{LSTM}^{(3)}$.
Although all four LSTM networks are active for this presence-pattern $\vec{p}_{t_{k}} = [1 , 1]$, this is not the case for the other presence-patterns. As an example, for the presence-pattern $\vec{p}_{t_{k}} = [1,0]$, only $\text{LSTM}^{(0)}$ and $\text{LSTM}^{(2)}$ generate output since  $P^\prime _{t_k} = \{0,2\}$. Note that $\text{LSTM}^{(0)}$ generates output at each time step since presence-pattern $[0,0]$ is included in the active set for any pattern, i.e., $[0,0] \in \forall \vec{P}_{t_{k}}$.

\textcolor{black}{From the mixture of experts perspective \cite{ozkan2014deterministic}, \cite{yuksel2012twenty}, \cite{kozat2010steady}, each LSTM network is an expert. 
The space of input vector sequences $\mathcal{X}$ is divided into regions in a hierarchical manner based on the presence-pattern at each time step. 
To generate the final estimate, we combine the outputs of the eligible experts, i.e., the active LSTM networks, as }
\begin{align}
\vec{\hat{h}}_{t_k} = \sum_{i=1}^{4} \alpha_{t_k}^{(i)}\bar{\vec{h}}_{t_k}^{(i)},
\end{align}
where $\alpha_{t_k}^{(i)}$ is the weight for the output of the $i^{\text{th}}$ LSTM network, i.e.,  $\text{LSTM}^{(i)}$, as learned in the following. To hold consistency in the time subscripts $t_k$, we represent the output of the $\text{LSTM}^{(i)}$ network with $\bar{\vec{h}}_{t_k}^{(i)}$ instead of $\vec{h}_{t_k}^{(i)}$, where $\bar{\vec{h}}_{t_k}^{(i)}$ is the most updated $\vec{h}_{t_k}^{(i)}$.
For example, in Fig. \ref{fig:model}, $\bar{\vec{h}}_{t_k}^{(0)} = \vec{h}_{t_{k-2}}^{(0)}$, $\bar{\vec{h}}_{t_k}^{(1)}$ and $\bar{\vec{h}}_{t_k}^{(3)}$ are the outputs of the $\text{LSTM}^{(1)}$ and $\text{LSTM}^{(3)}$ networks generated by using the input $\vec{x}_{m\Delta}$. Similarly, $\bar{\vec{h}}_{t_k}^{(2)}$ is the output of the $\text{LSTM}^{(2)}$ network generated with the input $\vec{x}_{(m-1)\Delta}$ since the last input for the $\text{LSTM}^{(2)}$ network is $\vec{x}_{(m-1)\Delta}$.
$\vec{\hat{h}}_{t_k}$ is the linear combination of these outputs. 

The number of the LSTM outputs to be combined varies with respect to the presence-pattern at each time step. We need an adaptive algorithm to determine weights $\alpha_{t_k}^{(i)}$ to hold combined outputs on the same scale. 
For this purpose we use $\mathrm{softmax}(\cdot)$ function to determine the combination weights $\alpha_{t_k}^{(i)}$, hence, the sum of the combination weights is always set to 1, i.e., $\sum_{i=1}^4 \alpha_{t_k}^{(i)} =1$. The combination weights are calculated as
\begin{equation}
\alpha_{t_k}^{(i)} = 
\begin{cases}
  \frac{\exp{\left( \vec{\tilde{w}}^{{(i)}^T}\vec{\tilde{h}}_{t_k}^{(i)} \right)}}
  {\sum_{j \in \vec{P}_{t_k}^\prime} \exp{\left(\vec{\tilde{w}}^{{(j)}^T}\vec{\tilde{h}}_{t_k}^{(j)}\right)}} 
  \qquad &\text{if} \hspace{0.2cm} i \in \vec{P}_{t_k}^\prime \\
  0	\qquad &\text{otherwise} 
\end{cases},
\label{eq:weigthL2}
\end{equation}
where $\vec{\tilde{w}}_{t_k}^{(i)} \in  \mathbb{R}^{4+q}$. $\vec{\tilde{h}}_{t_k}^{(i)} \in \mathbb{R}^{4+q}$ is defined as $\vec{\tilde{h}}_{t_k}^{(i)} = [\vec{p}_{t_{k}}; \vec{p}^{(i)}; \vec{\bar{h}}_{t_k}^{(i)}]$, i.e., we incorporate the missingness information in length-2 window to the weight calculations by appending the presence-patterns of the current input and the $\text{LSTM}^{(i)}$ network to the LSTM network outputs. Note that we consider only the outputs of the LSTM networks in the active set $\vec{P}^\prime_{t_k}$ to calculate combined output $\vec{\hat{h}}_{t_k}$. The final estimate of the desired signal is calculated by
\begin{align}
\hat{d}_{t_k} = \vec{\hat{w}}_{t_k}^T\vec{\hat{h}}_{t_k}.
\end{align}

In Section \ref{subsec:Generic}, we explain the proposed architecture for the generic case, i.e., length-$L$ window.


\subsection{Generic Tree-LSTM Architecture}\label{subsec:Generic}
In this subsection, we consider the Tree-LSTM architecture for the generic case, i.e., the depth of the Tree-LSTM is $L$. 
The architecture contains $2^L$ different LSTM networks, where each LSTM network specializes in estimation of desired signal $d_{t_k}$ from an input sequence $[\vec{x}_{t_k}, \ldots, \vec{x}_{t_1}]$ with a particular presence-pattern $\vec{p}_{t_k}\in \mathbb{R}^L$. 
$\vec{W}_z^{(j)}$, $\vec{W}_i^{(j)}$, $\vec{W}_f^{(j)}$, $\vec{W}_o^{(j)} \in \mathbb{R}^{q\times m}$ are the input weight matrices and $\vec{R}_z^{(j)}$, $\vec{R}_i^{(j)}$, $\vec{R}_f^{(j)}$, $\vec{R}_o^{(j)} \in \mathbb{R}^{q\times q}$ are the recurrent weight matrices of the $\text{LSTM}^{(j)}$ network.

\renewcommand{\thealgorithm}{}

\begin{algorithm}
    \caption{The Tree-LSTM Network Regressor}\label{alg:Algorithm1}
    \begin{algorithmic}[1]
    	\STATE $k = 0$
    	\FOR {m = 1 \TO $L$}
    		\STATE $\vec{x}_{t_k} = \vec{x}_{m\Delta}$
    	\ENDFOR
        \FOR{$m = L+1$ \TO $N$}
        	\IF{$\vec{x}_{(m)\Delta}$ exists}
        		\STATE $k = k + 1$
        		\STATE $\vec{x}_{t_k} = \vec{x}_{m\Delta}$
        	\ENDIF
        	\IF{$\vec{x}_{(m-L)\Delta}$ exists}
        		
        		\STATE $ \vec{h}_{t_{k-L}}^{(0)}, \vec{c}_{t_{k-L}}^{(0)}\Leftarrow \text{LSTM}^{(0)}(\vec{x}_{t_{k-L}}) $
        	\ELSE 
        		\STATE $ \vec{h}_{t_{k-L}}^{(0)} = \vec{h}_{t_{k-L-1}}^{(0)}$
        		\STATE $\vec{c}_{t_{k-L}}^{(0)} = \vec{c}_{t_{k-L-1}}^{(0)}$
        	\ENDIF
        	\STATE $\vec{\bar{h}}^{(0)}_{t_k} = \vec{h}_{t_{k-L}}^{(0)}$

            \FORALL{$i \in \vec{S}^{\prime}_{{t_k}}$}     	
                \STATE $\vec{H}_{t_{k},1}^{(i)} = \vec{h}_{t_{k-L-1}}^{(0)}$
                \STATE $\vec{C}_{t_{k},1}^{(i)} = \vec{c}_{t_{k-L-1}}^{(0)}$
                \FOR{$j = 1$ \TO $||\vec{p}^{(i)}||_1$} 
                	\STATE $ \vec{H}_{t_{k},j}^{(i)}, \vec{C}_{t_{k},j}^{(i)}\Leftarrow \text{LSTM}^{(i)}(\vec{X}^{(i)}_{t_k,j})$ 
                \ENDFOR      
                \STATE $\vec{\bar{h}}_{t_k}^{(i)} = \vec{{H}}_{t_{k}, ||\vec{p}^{(i)}||_1}^{(i)}$              
            \ENDFOR

            \FORALL{$i \in \vec{P}^{\prime}_{t_k}$}
            	\STATE $\vec{\tilde{h}}^{(i)}_{t_k} = [\vec{p}_{t_k};\vec{p}^{(i)};\vec{\bar{h}}^{(i)}_{t_k}]$
                \STATE $\alpha^{(i)} = \text{softmax}(\vec{\tilde{w}}^{(i)^T}\vec{\tilde{h}}^{(i)}_{t_k})$
            \ENDFOR
            
            \STATE $\vec{\hat{h}_{t_k}} = \sum_{i \in \vec{P}^{\prime}}\alpha^{(i)}\vec{\bar{h}}^{(i)}_{t_k}$
            
			\STATE $d_{t_k} = \vec{\hat{w}}^T \vec{\hat{h}_{t_k}}$            
            
            \STATE $e_{t_k} = \frac{1}{2} (d_t - \hat{d}_t)^2$      
        \ENDFOR
    \end{algorithmic}
\end{algorithm}

To generate the estimate of the desired signal $d_{t_k}$, we first create the presence-pattern $\vec{p}_{t_k}$ and its corresponding sets $\vec{P}_{t_k}$, $\vec{P}^{\prime}_{t_k}$ by considering the existence of the last $L$ input vectors, i.e., $[\vec{x}_{(m-L+1)\Delta}, \ldots, \vec{x}_{m\Delta}]$, where $m\Delta = t_k$. Based on this presence-pattern, we choose $2^{||\vec{p}_{t_k}||_{1}}$ LSTM networks, i.e., $\text{LSTM}^{(j)}$, where $j \in \vec{P}^{\prime}$, among the total $2^L$ LSTM networks in the architecture. As described in Algorithm, firstly, the main LSTM network, $\text{LSTM}^{(0)}$, generates its state and output vectors, $ \vec{h}_{t_{k-L}}^{(0)}, \vec{c}_{t_{k-L}}^{(0)}$, by processing the input $\vec{x}_{(m-L)\Delta}$, if it exists. Otherwise, we directly use the previous state and output vectors of the main LSTM network. We then pass these state and output vectors of the main LSTM network, i.e., $\vec{c}^{(0)}_{t_{k-L}}$ and $\vec{h}^{(0)}_{t_{k-L}}$, to the leaf LSTM networks, i.e., $\text{LSTM}^{(j)}$, where $j \in \vec{P}^{\prime}_{t_k}$, as their initial state and recurrent input vectors. Each active leaf LSTM network processes its corresponding length-$||\vec{p}^{(i)}||_1$ input sequence $\vec{X}^{(i)}_{t_k}$ and generates its output vector. Note that in Algorithm, $\vec{H}_{t_{k}}^{(i)}$, $\vec{C}_{t_{k}}^{(i)}\in \mathbb{R}^{q\times||\vec{p}^{(i)}||_1}$ are the matrices storing the state and output vectors of the $\text{LSTM}^{(i)}$ in their columns, respectively.
To simplify the notation, we denote the last output vector of each LSTM network by $\vec{\bar{h}}^{(i)}_{t_k}$. 
We then create $\vec{\tilde{h}}^{(i)}_{t_k}$ vectors for our combination algorithm by appending $\vec{p}_{t_k}$ and $\vec{p}^{(i)}$, which represents the presence-pattern of the $\text{LSTM}^{(i)}$ network, to these output vectors, i.e., $[\vec{p}_{t_{k}}; \vec{p}^{(i)}; \vec{\bar{h}}_{t_k}^{(i)}]$. 
Here, each combination weight is conditioned on the input presence-pattern and the assigned presence- pattern of the LSTM network. 
We generate the combination weights as
\begin{equation}
\alpha_{t_k}^{(i)} = 
\begin{cases}
  \frac{\exp{\left(\vec{\tilde{w}}^{{(i)}^T}\vec{\tilde{h}}_{t_k}^{(i)}\right)}}
{\sum_{j \in \vec{P}_{t_k}^{\prime}} \exp{\left(\vec{\tilde{w}}^{{(j)}^T}
\vec{\tilde{h}}_{t_k}^{(j)}\right)}} \qquad &\text{if} \hspace{0.2cm} i \in  \vec{P}_{t_k}^{\prime} \\
  0	\qquad &\text{otherwise} 
\end{cases},
\label{eq:weigthL}
\end{equation}
where $\vec{\tilde{w}}^{(i)} \in \mathbb{R}^{q+2L}$. We use $\alpha ^{(i)}_{t_k} \in \mathbb{R}$ to linearly combine the outputs of the LSTM networks as
\begin{align}
\vec{\hat{h}}_{t_k} = \sum_{i \in \vec{P}^{\prime}} \alpha_{t_k}^{(i)}\bar{\vec{h}}_{t_k}^{(i)},
\end{align}
where $\vec{\hat{h}}_{t_k} \in \mathbb{R}^{q}$ is the final output vector of the architecture. Finally, we generate the estimate of the desired signal by
\begin{align}
\hat{d}_{t_k} = \vec{\hat{w}}_{t_k}^T\vec{\hat{h}}_{t_k},
\end{align}
where, $\vec{\hat{w}}_{t_k} \in \mathbb{R}^{q+1}$ is the final regression weights.

\remark{
We introduce the most extensive variant of the Tree-LSTM architecture, i.e., all of the $2^L$ LSTM networks are included. Since we provide an adaptive combination algorithm working on any number of LSTM networks, one can use only a set of desired LSTM networks by exclusively altering the set $\vec{P}_{t_k}$. As an example, for the length-3 presence-pattern $\vec{p}_{t_k} = [1,1,1]^T$, one can employ only the LSTM networks with the presence-patterns $[1,0,0]^T$, $[1,1,0]^T$ and $[1,1,1]^T$ instead of the $2^3 = 8$ LSTM networks, which corresponds to the combination of $1-, 2-$ and $3-$step ahead predictors.  Therefore, the number of LSTM networks can be reduced to avoid overfitting issues and accelerate the training of the architecture thanks to our adaptive combination algorithm. In addition, one can also use a common weight vector $\vec{\tilde{w}}$ to calculate $\alpha _{t_k} ^{(i)}$'s for all of the LSTM networks instead of assigning a unique weight vector $\vec{\tilde{w}}^{(i)}$ for each of them.
}

\remark
The complexity of the new architecture is in
the same order of the complexity of the conventional LSTM
architectures in terms of the sequence length, i.e., $N$. 
In Table \ref{tab:table1}, we provide the computational loads
in terms of the number of required multiplication operations to process a length-$N$ sequence containing $M$ missing samples for the Tree-LSTM architecture and the conventional algorithms. In Table \ref{tab:table1}, LSTM-ZI is the network that imputes all-zero vectors for the missing inputs. LSTM-FI represents the LSTM network using forward-filling imputation technique together with a binary missingness indicator as another feature. 
In the vanilla LSTM architecture, there exists four matrix-vector multiplications for the input, i.e., $\vec{W}\vec{x}_{t_k}$, four matrix-vector multiplications for the recurrent input, i.e., $\vec{R}\vec{h}_{t_{k-1}}$, and three vector-vector multiplications between the gates, i.e., \eqref{eq:ct} and \eqref{eq:ht}, which correspond to $4q^2 + 4qm + 3q$ multiplication operations in total.
Since the LSTM-FI algorithm extends the feature vector with a binary missingness indicator, the input size is $m+1$ for this algorithm, which requires $4$ additional multiplication operations, i.e., $4q^2 + 4qm + 7q$.
Since the LSTM-ZI and LSTM-FI algorithms impute the missing inputs with the all-zero vectors and the previous existing input vectors, respectively, these algorithms require $N$ LSTM operations to process a length-$N$ sequence. 
On the other hand, the Tree-LSTM architecture processes only the existing inputs in the sequence, which requires $N-M$ Tree-LSTM network operations to process the same sequence. 
For each step of the Tree-LSTM architecture, the main LSTM network processes the each existing input once, however, the number of active leaf LSTM alters based on the presence-pattern of the input. 
For a length-$L$ window, there exist $2^L$ different leaf LSTM networks, which require $2^L \times \frac{L}{2}$ LSTM operations in total if all $L$ inputs in the window exist. 
If one input is missing, the total number of LSTM operations for this window decreases to $2^{L-1} \times \frac{L-1}{2}$. 
Similarly, the total number of LSTM operations for this window is $2^{L-2} \times \frac{L-2}{2}$ for the case two inputs are missing. 
We emphasize that the computational load of the Tree-LSTM architecture depends on the distribution of the missing inputs in addition to number of missing inputs. In the worst (also unrealistic) case, the missing and existing inputs are completely separated, the total computational load for our architecture is $(N-M)(1 + {2^{L-1}L})(4q^2 + 4qm + 3q)$. However, while the distribution of the missing inputs goes to the uniform distribution, the computational load our algorithm rapidly decreases and converges to $(N-M)(1 + \frac{2^{L(1-r)}L(1-r)}{2})(4q^2 + 4qm + 3q)$, where $r =\frac{M}{N}$ is the missingness ratio.
We point out that the computational load in terms of the number of required multiplication operations for our algorithm decreases as the ratio of missing inputs increases. 
In particular, our architecture is more efficient than the conventional architectures for high values of missingness ratio, $r$, and small values of the window length $L$. For example, the computational load for the Tree-LSTM architecture is less than the computational load for the LSTM-ZI architecture when $i)$ $r > 0.5$ if $L =2$, $ii)$ $r > 0.60$ if $L =3$ and $iii)$ $r > 0.65$ if $L =4$ in the optimal case, i.e., the missing inputs are far from each other as much as possible. Since the computational load for the LSTM-FI architecture is higher than the computational load for the LSTM-ZI architecture, our algorithm is also more efficient than LSTM-FI architecture for these parameters.

\begin{table}
 \centering
 \resizebox{\columnwidth}{!}{
  \begin{tabular}{|l||*{2}{c|}} \hline
  \makebox{ Architecture}
  &\makebox{\hspace{.60cm} Computational Load \hspace{.60cm} } \\\hline\hline
\rule{0pt}{0.3cm}
  \hspace{.05cm} LSTM-ZI & $N(4q^2 + 4qm + 3q)$ \\\hline
  \rule{0pt}{0.3cm}
  \hspace{.05cm} LSTM-FI  & $N(4q^2 + 4qm + 7q)$ \\\hline
  \rule{0pt}{0.3cm}
  \hspace{.05cm} Tree-LSTM (max)  & $(N-M)(1 + {2^{L-1}L})(4q^2 + 4qm + 3q)$ \vspace{0.02cm} \\\hline
  \rule{0pt}{0.3cm}
  \hspace{.05cm} Tree-LSTM (min)& $(N-M)(1 + \frac{2^{L(1-r)}L(1-r)}{2})(4q^2 + 4qm + 3q)$ \vspace{0.02cm} \\\hline
   \end{tabular}
  }
  \caption{The number of multiplication operations in the forward pass of the LSTM-ZI, LSTM-FI and the Tree-LSTM architectures to process a sequence with length-$N$, where $M$ is the number of missing inputs. LSTM-ZI is the network that imputes all-zero vectors for the missing inputs. LSTM-FI represents the LSTM network using forward-filling method and a binary missingness indicator as another feature. Tree-LSTM (max) and Tree-LSTM (min) are the maximum and the minimum computational loads for our architecture. }\label{tab:table1}
  \vspace{-0.3cm}
\end{table}

\remark{We point out that our Tree-LSTM architecture can be straightforwardly extended as follows. $i)$ One can combine multiple Tree-LSTM networks with distinct window lengths in the sense of combination of mixture of experts. $ii)$ One can start with a small window length e.g., $L = 1$, and then increase this window length as the leaf LSTM networks are trained. We note that these trained leaf LSTM networks constitute half of the leaf LSTM networks when we increase the window length by $1$, i.e., the leaf LSTM networks containing a $0$ in the first entry of their presence-patterns. By this way, the capacity of the Tree-LSTM architecture incrementally grows as we increase the window length. In addition, since half of the the leaf LSTM networks starts as substantially trained weights, the trainability characteristics of the architecture for the large window lengths may increase.
}

\section{Simulations} \label{sec:Sim}
	In this section, we illustrate the regression performance of the proposed Tree-LSTM architecture under different scenarios with respect to the state-of-the-art algorithms in various real-life datasets. In the first part, we focus on the next value prediction problem over various financial datasets such as the New York stock exchange (NYSE) \cite{NYSE} and the Bitcoin \cite{BTC}. In the second part, we compare our algorithm with the other architectures on the several real life datasets such as kinematics \cite{ltorgo} and California housing \cite{delve}. We also illustrate the performance of our architecture in underfitting and overfitting (in terms of the depth of the tree) scenarios. 
	
Throughout this section, "TL" represents the Tree-LSTM architecture. In the first two part, we use the Tree-LSTM architecture with the depth $L = 3$ to have sufficient length input sequences for the leaf LSTM networks by keeping the number of the LSTM networks in a certain limit. The single LSTM network using $i)$ the zero imputation and $ii)$ the forward-filling imputation with a missingness indicator algorithms are denoted by "ZI" and "FI", respectively.

Since the datasets used in our simulations do not have separate training and test sets, we split the sequences in each dataset such that the first $\%60$ of the sequence is used for training and the remaining $\%40$ is for test. 
We also insert missingness to these sets by randomly deleting certain inputs.
To evaluate the performance of the algorithms with respect to the different missingness ratios we generate two different sets from each dataset such that randomly selected $\%30$ and $\%70$ of the sequences are missing. In the simulations, these datasets with $\%30$ and $\%70$ missingness ratios are represented by "-F" (frequent) and "-S" (sparse), respectively.
For the training of the networks, we employ Stochastic Gradient Descent (SGD) algorithm \cite{jaeger2002tutorial} with a constant learning rate. 
We use 5-fold cross validation for the parameter selection. 

\begin{figure}[t]
  \centering
  \includegraphics[width=.50\textwidth]{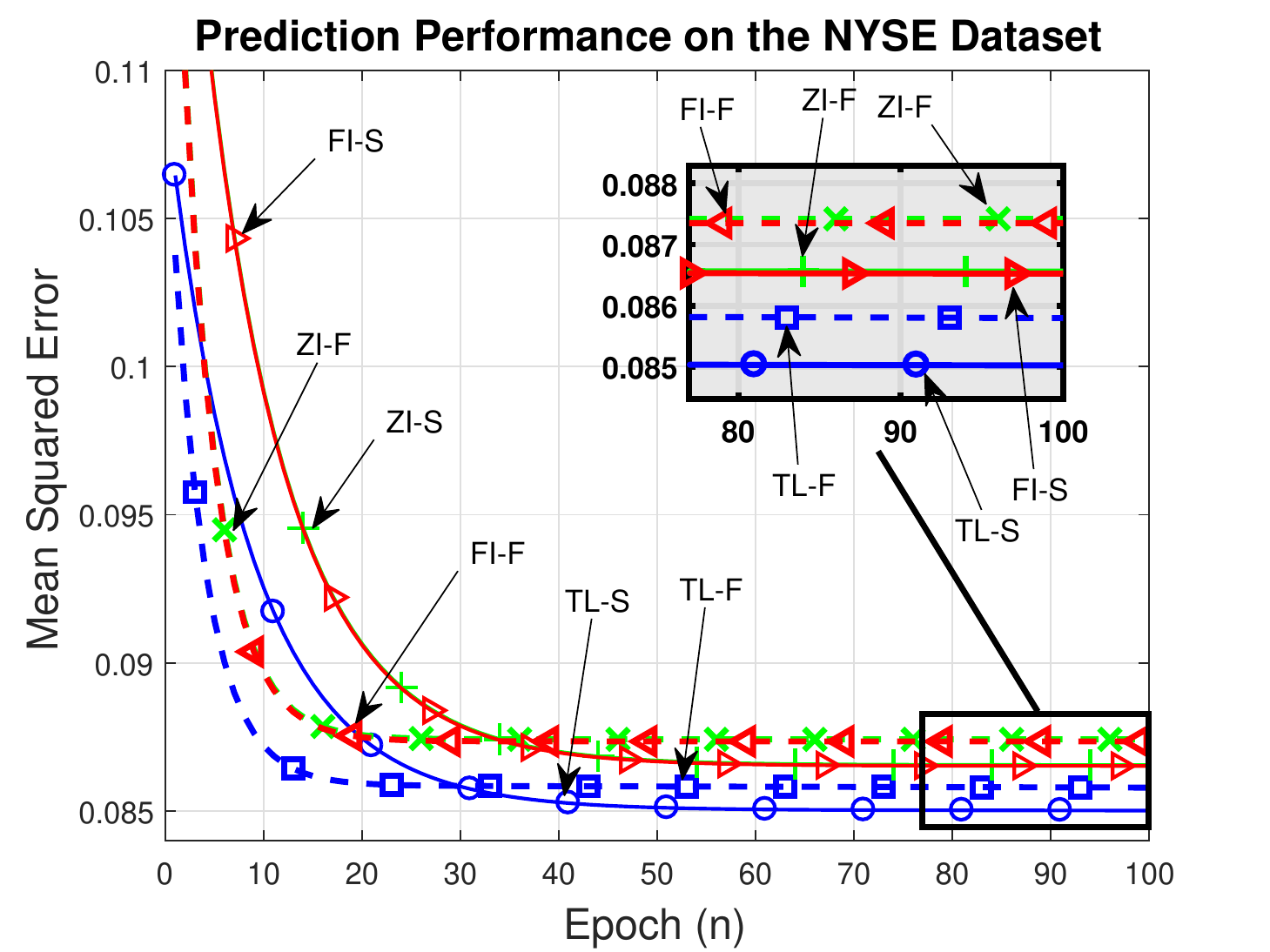}\\
  \caption{Prediction performances for the New York Stock Exchange dataset.}\label{fig:NYSE}
  \vspace{-0.3cm}
\end{figure}

\subsection{Financial Datasets} \label{sec:Financial}
	In this subsection, we evaluate the performances of the Tree-LSTM architectures and the single LSTM architectures employing the zero imputation and the forward-filling techniques. The LSTM network with the forward-filling algorithm uses the existence of the inputs as another feature in the input vectors. Therefore, for a dataset with the input size $m$, this algorithm has the input size $m+1$.
	
	We first evaluate the performances of the algorithms on NYSE dataset. The dataset contains the stock prices of 36 different companies over 5651 days (22 years), where we randomly select the third company, i.e., Amer-Brands, for the simulations. For this data, the input is scalar $x_{t_k} \in \mathbb{R}$, i.e., the input size $m = 1$, and the desired output $d_{t_k} \in \mathbb{R}$, where $d_{t_k} = x_{t_{k+1}}$. 
For the parameter selection, we make a grid search on the number of hidden neurons and the learning rate in the intervals $q = [3, 10]$ and $\eta = [10^{-1}, 10^{-5}]$, respectively. 
We choose the number of hidden neurons as $q = 10$ and the learning rate as $10^{-3}$ using cross validation. All of the weights in the networks are initiated from the Gaussian distribution $\mathcal{N}(0,10^{-2})$.
	
	In Fig. \ref{fig:NYSE}, we illustrate the prediction performance of the algorithms in terms of the mean squared error on the test set per epoch. For both experiments conducted with different missing rates, the Tree-LSTM architecture significantly outperforms the other two architectures in terms of the steady-state performance, thanks to its structure providing a unique  response to each presence-pattern. The forward-filling imputation with an existence indicator algorithm is slightly better than the zero imputation algorithm in terms of the steady-state error. The Tree-LSTM architecture has also a faster convergence rate compared to the other LSTM architectures. These results show that assigning a unique LSTM network for each presence-pattern and then combining their outputs successfully models the effect of missing data. Our Tree-LSTM architecture outperforms the state-of-the-art architectures in terms of both convergence rate and the steady-state performance in this one-step ahead estimation task.

\begin{figure}[t]
  \centering
  \includegraphics[width=.50\textwidth]{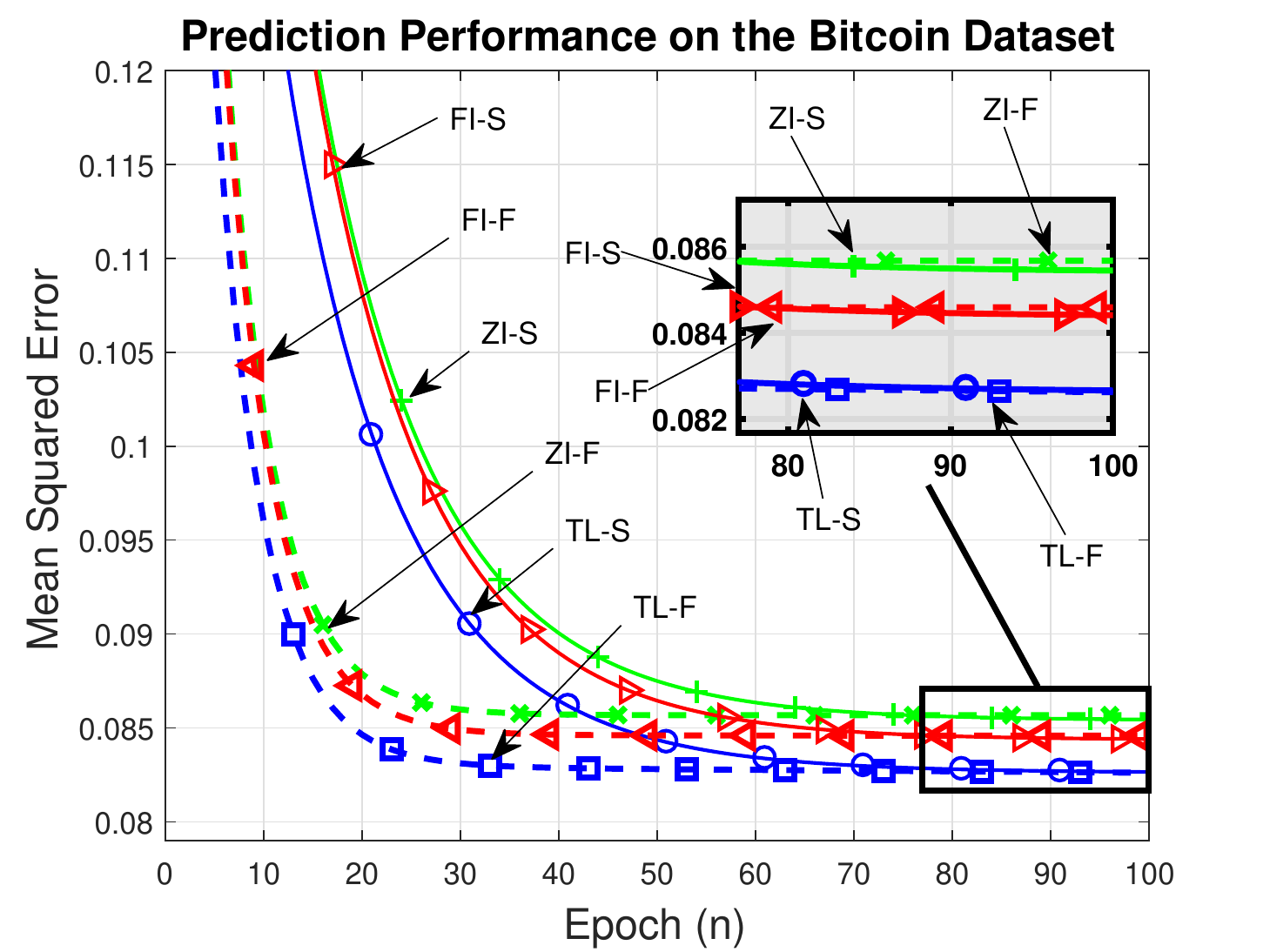}\\
  \caption{Prediction performances for the Bitcoin dataset.}\label{fig:BTC}
  \vspace{-0.3cm}
\end{figure}

We also test our algorithm in Bitcoin \cite{BTC} dataset, which is a more challenging and unstable data compared to NYSE. The dataset contains the price of Bitcoin in terms of USD. Similar to NYSE dataset, the input is scalar $x_{t_k} \in \mathbb{R}$, i.e., the input size $m = 1$, and the desired output $d_{t_k} \in \mathbb{R}$, where $d_{t_k} = x_{t_{k+1}}$. For the parameter selection, we make a grid search on the number of hidden neurons and the learning rate in the intervals $q = [3, 10]$ and $\eta = [10^{-1}, 10^{-5}]$, respectively. 
We choose the number of hidden neurons as $q = 10$ and the learning rate as $10^{-4}$ using fivefold cross-validation. We initiate the weights from the distribution $\mathcal{N}(0,10^{-2})$.

	In Fig. \ref{fig:BTC}, we demonstrate the prediction performance of the algorithms in terms of the mean squared error on the test set per epoch. For both missing rates, our architecture has a better steady-state performance compared to the other two algorithms. In terms of the convergence rate, all of the architectures have similar performances in this dataset. The results show that the Tree-LSTM architecture significantly outperforms the other algorithms thanks to its novel structure, which separately processes each pattern and adaptively combines them by incorporating the missingness information.

\begin{figure}[t]
  \centering
  \includegraphics[width=.50\textwidth]{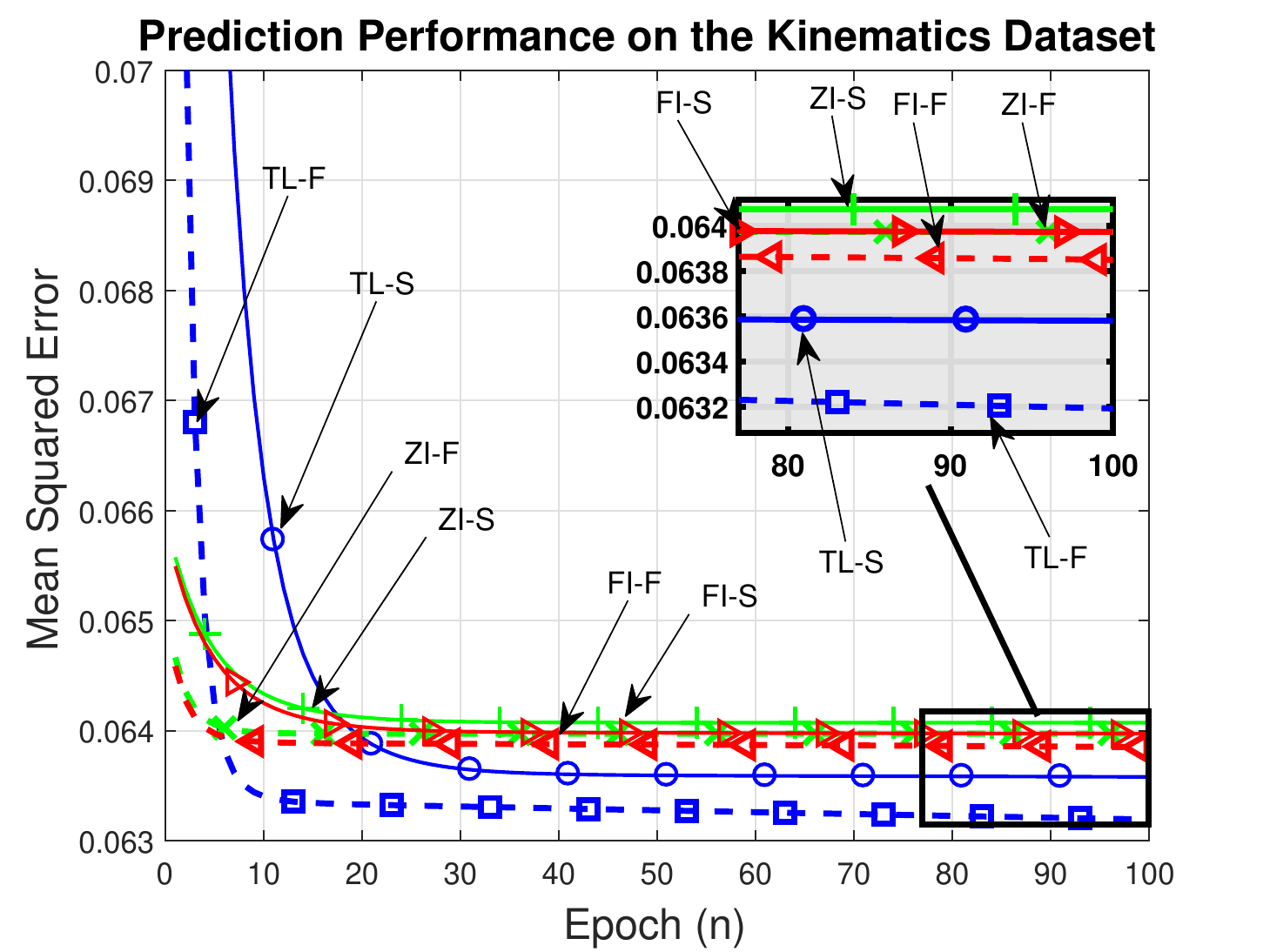}\\
  \caption{Regression performances for the Kinematics dataset.}\label{fig:Kinematics}
  \vspace{-0.3cm}
\end{figure}

\subsection{Real Life Datasets} \label{sec:RealLife}
	In this subsection, we compare the performances of the algorithms over several real life datasets under different missing rates. The LSTM network with the forward-filling algorithm uses the existence of the inputs as another feature in the input vectors. Therefore, for a dataset with the input size $m$, this algorithm has the input size $m+1$.
	We test our algorithms on kinematics \cite{ltorgo} and California housing \cite{delve} datasets. These datasets contain an input vector sequence and the corresponding desired signal for each time step.
\begin{itemize}
	\item Kinematic dataset is a simulation of 8-link all-revolute robotic arm, where the aim is to predict the distance of the effector from the target. The original input vector size $m = 8$ and we set the number of hidden neurons $q = 8$ for both LSTM and TG-LSTM networks. For the SGD algorithm, we select the constant learning rate $\eta = 10^{-4}$ from the interval $[10^{-5}, 10^{-2}]$ using the cross-validation.

	\item California Housing dataset contains the house prices in the California area and the aim is to estimate the median of these house prices. The input vector $\vec{x}_{t_k} \in \mathbb{R}^{8}$. We set the number of hidden neurons $q = 8$, and the constant learning rate $\eta = 10^{-4}$ from the interval $[10^{-5}, 10^{-2}]$.
\end{itemize}

\begin{figure}[t]
  \centering
  \includegraphics[width=.50\textwidth]{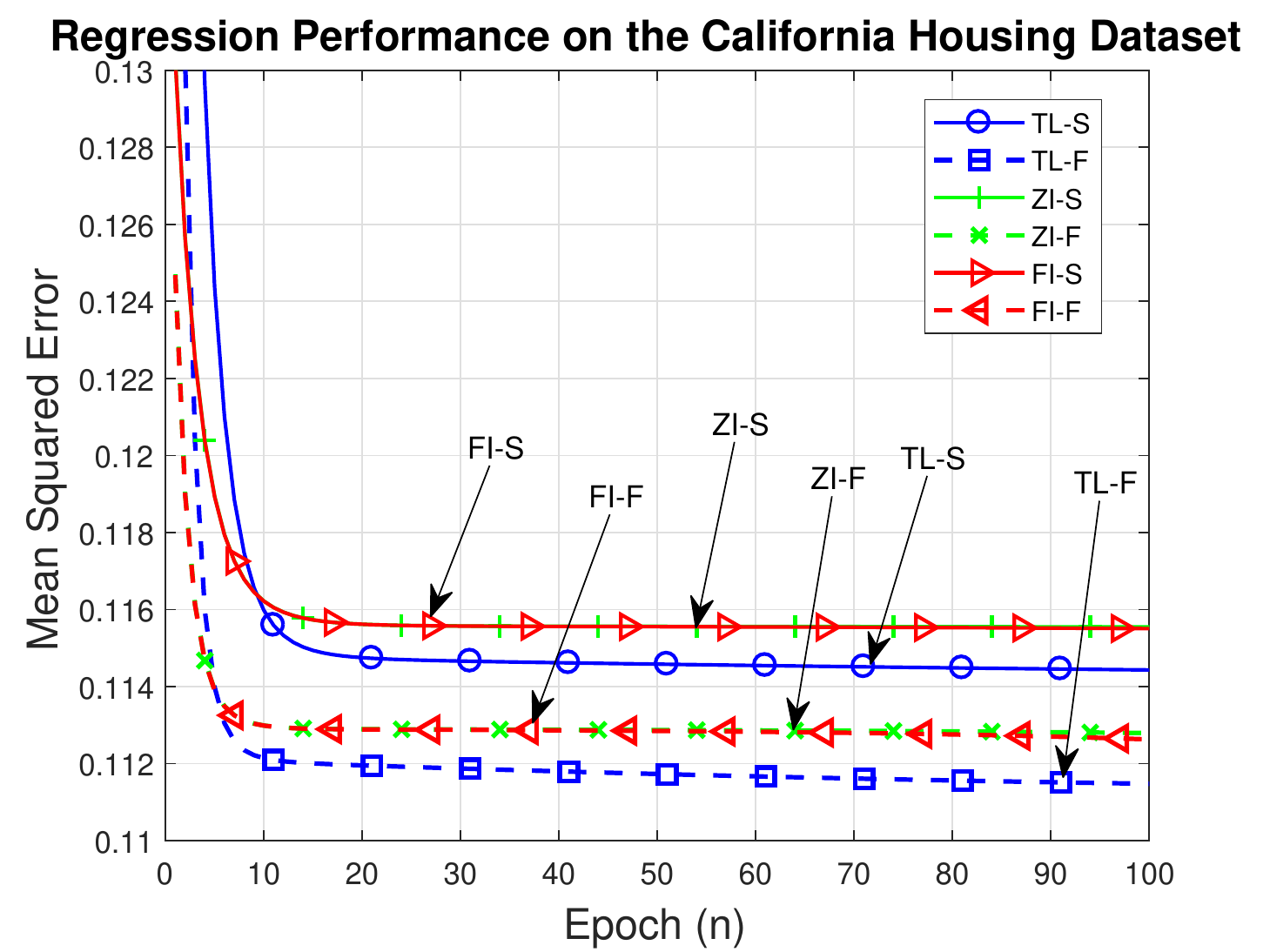}\\
  \caption{Regression performances for the California housing dataset.}\label{fig:Cal_Housing}
  \vspace{-0.3cm}
\end{figure}

In Fig. \ref{fig:Kinematics} and Fig. \ref{fig:Cal_Housing}, we illustrate the regression performance of the algorithms in terms of the mean squared error per epoch for kinematics and California housing datasets, respectively. In these simulations, the Tree-LSTM architecture captures the sequential pattern of the data with a faster convergence rate compared to the other two algorithms. The LSTM architecture using forward-filling imputation method is slightly better than the LSTM architecture using the zero imputation technique in terms of the steady-state error. However, our architecture significantly outperforms the other two methods in terms of the steady-state error. These results show that our algorithm successfully handles the effect of missing samples and models the underlying structure by using only the existing data compared to the other methods.

\begin{figure}[t]
  \centering
  \includegraphics[width=.50\textwidth]{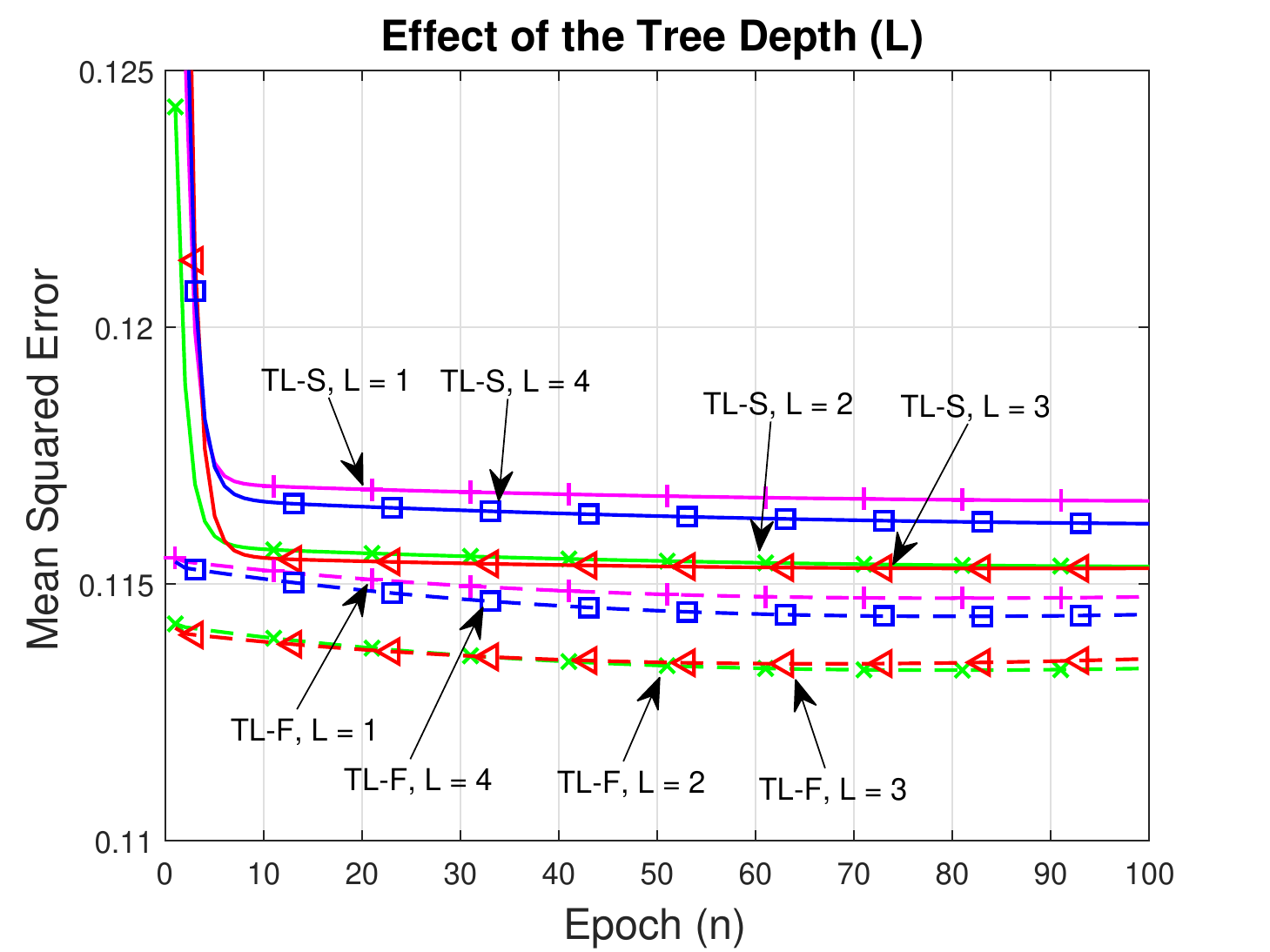}\\
  \caption{The effect of the tree depth on the regression performance the Tree-LSTM architecture for the California housing dataset.}\label{fig:Comp}
  \vspace{-0.3cm}
\end{figure}

In Fig. \ref{fig:Comp}, we illustrate the effect of the parameter tree depth $L$ on the performance of our architecture over the California housing dataset. For this simulation, we use $\{1,2,3,4\}$ as the depth of the tree, the number of hidden neurons $q = 8$, and the learning rate $\eta = 10^{-4}$. When we compare the performance of the algorithm with the same depth for different missing rates, the steady-state performance on the frequent data is significantly higher than the performance on the sparse data. For the same missing rates and different depth of the tree, the Tree-LSTM architecture with the depths $ L = 2$ and $L = 3$ achieve the highest performances in terms of the steady-state performance. These two networks outperform even the Tree-LSTM architecture with the depth $L=4$ since the size of the data is not sufficient for learning to combine the outputs of 16 LSTM networks. Here, we observe modelling capacity vs. trainability trade-off \cite{collins2016capacity} of the recurrent neural networks, i.e., while the number of the parameters of an RNN increases, its potential modelling capability increases as well, however, the training process becomes more difficult and the RNN may not achieve its potential steady-state performance.  Therefore, $L=2$ and $L=3$ are the optimal choices for this dataset in terms of the steady-state performance.

\section{Conclusion} \label{sec:Conc}
We have studied nonlinear regression of variable length sequential data suffering from missing samples in a sequential setting and introduce a novel architecture based on the LSTM network, namely, the Tree-LSTM network. 
In the Tree-LSTM architecture, we use one main LSTM network and a certain number of leaf LSTM networks, where each LSTM network is an expert from the perspective of mixture of experts. Each LSTM network is responsible for processing the data sequence with a particular presence-pattern, i.e., we divide the input space into the regions based on the missingness information in a hierarchical manner.
We adaptively combine the outputs of these LSTM networks based on the presence pattern and generate the final output at each time step. Here, only experts eligible to process the received input sequence contribute to the final output.
In our architecture, we incorporate the missingness information by selecting the particular leaf LSTM networks based on the missingness pattern of the input sequence. 
Furthermore, in terms of the number of multiplication operations the computational load of our algorithm is less than the computational load of the conventional algorithm under high number of missing input values.
We also point out that our architecture can be straightforwardly applied to the other similar networks such as GRU. 
The introduced algorithm protects the model against the deteriorations since it avoids $i)$ inclusion of arbitrarily generated and unrelated inputs, $ii)$ multiple imputations of the same input and also $iii)$ unreliable assumptions on the missing data, since our architecture uses only the existing inputs without any assumption on the missing data. 
We demonstrate significant performance improvements achieved by the introduced architecture with respect to the state-of-the-art methods in several different datasets.

\bibliographystyle{IEEEtran}
\bibliography{my_references}

\vfill

\end{document}